\documentclass[10pt,twocolumn,letterpaper]{article}

\usepackage{iccv}
\usepackage{times}
\usepackage{epsfig}
\usepackage{graphicx}
\usepackage{amsmath}
\usepackage{amssymb}
\usepackage{multirow}
\usepackage{bm}
\usepackage{epstopdf}

\usepackage[breaklinks=true,bookmarks=false]{hyperref}

\iccvfinalcopy % *** Uncomment this line for the final submission

\def\iccvPaperID{****} % *** Enter the ICCV Paper ID here

\ificcvfinal\fi

\begin{document}

%%%%%%%%% TITLE
\title{Transferable Contrastive Network for Generalized Zero-Shot Learning}

\author{Huajie Jiang$^{1,2,3,4}$, Ruiping Wang$^{1,2}$, Shiguang Shan$^{1,2}$, Xilin Chen$^{1,2}$\\
$^{1}$ Key Laboratory of Intelligent Information Processing of Chinese Academy of Sciences (CAS),\\
Institute of Computing Technology, CAS, Beijing, 100190, China\\
$^{2}$ University of Chinese Academy of Sciences, Beijing, 100049, China\\
$^3$Shanghai Institute of Microsystem and Information Technology, CAS, Shanghai, 200050, China\\
$^4$School of Information Science and Technology, ShanghaiTech University, Shanghai, 200031, China\\
{\tt\small huajie.jiang@vipl.ict.ac.cn, \{wangruiping, sgshan, xlchen\}@ict.ac.cn}
}

\maketitle
\ificcvfinal\thispagestyle{empty}\fi

%%%%%%%%% ABSTRACT
\begin{abstract}
Zero-shot learning (ZSL) is a challenging problem that aims to recognize the target categories without seen data, where semantic information is leveraged to transfer knowledge from some source classes. Although ZSL has made great progress in recent years, most existing approaches are easy to overfit the sources classes in generalized zero-shot learning (GZSL) task, which indicates that they learn little knowledge about target classes. To tackle such problem,  we propose a novel Transferable Contrastive Network (TCN) that explicitly transfers knowledge from the source classes to the target classes. It automatically contrasts one image with different classes to judge whether they are consistent or not. By exploiting the class similarities to make knowledge transfer from source images to similar target classes, our approach is more robust to recognize the target images. Experiments on five benchmark datasets show the superiority of our approach for GZSL.
\end{abstract}

%%%%%%%%% BODY TEXT
\section{Introduction}

Object recognition is one of the basic issues in computer vision. It has made great progress in recent years with the rapid development of deep learning approaches \cite{Krizhevsky2012,Christian2015,Simonyan2014,Kaiming2016}, where large numbers of labeled images are required, such as ImageNet \cite{ILSVRC15}. However, collecting and annotating large numbers of images are difficult, especially for fine-grained categories in specific domains. Moreover, such supervised learning approaches can only recognize a fixed number of categories, which is not flexible. In contrast, humans can learn from only a few samples or even recognize unseen objects. Therefore, learning visual classifiers with no need of human annotation is becoming a hot topic in recent years.

Zero-shot learning (ZSL) aims to learn classifiers for the target categories where no labeled images are accessible. It is accomplished by transferring knowledge from the source categories with the help of semantic information. Semantic information can build up the relations among different classes thus to enable knowledge transfer from source classes to target classes. Currently the most widely used semantic information includes attributes \cite{lampert2009learning,farhadi2009describing} and word vectors \cite{frome2013devise,akata2015evaluation}. Traditional ZSL approaches usually learn universal visual-semantic transformations among the source classes and then apply them to the target classes. In this way, the visual samples and class semantics can be projected into a common space, where zero-shot recognition is conducted by the nearest neighbor approach.

\begin{figure}[t]
\centering
\includegraphics[height=4.5cm]{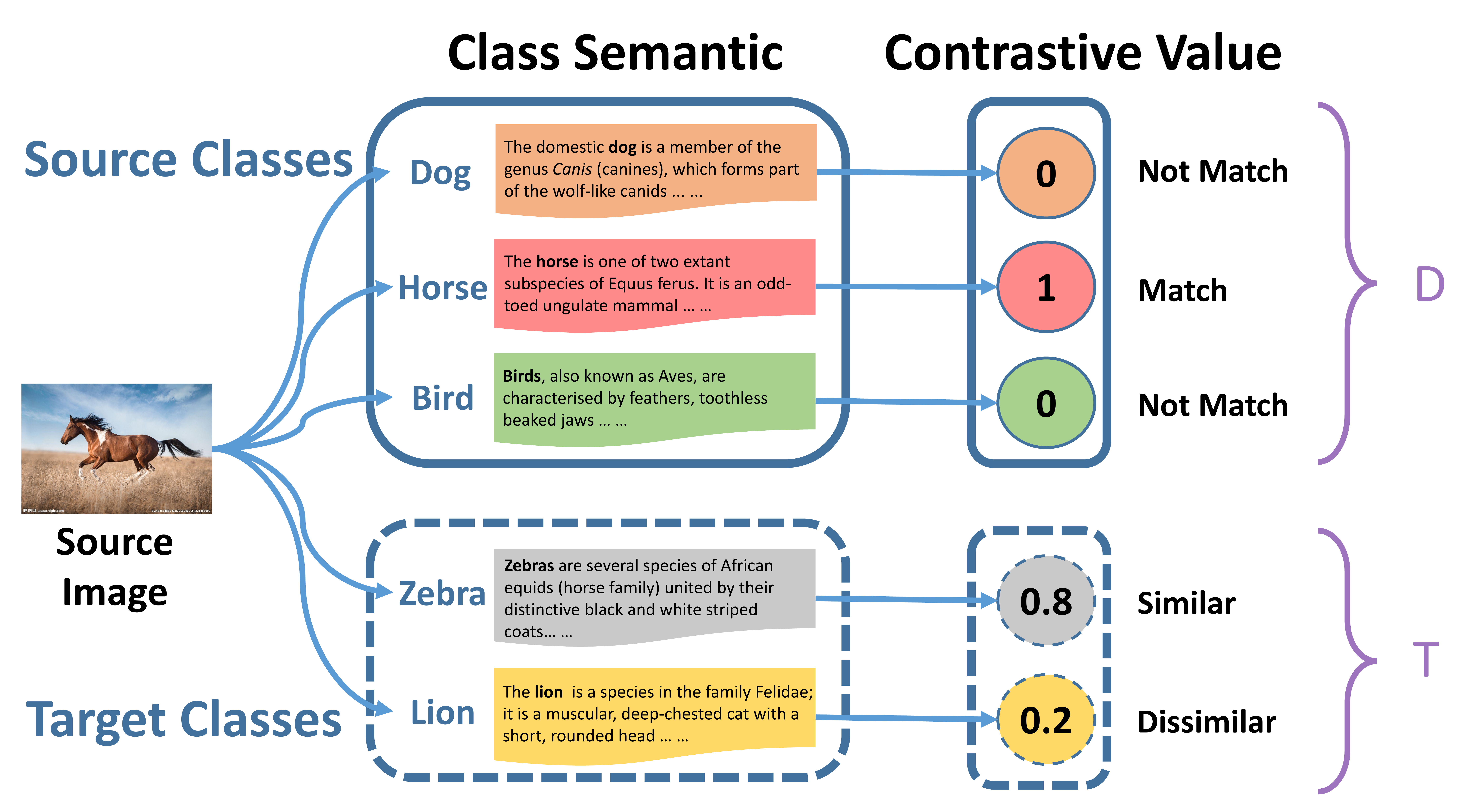}
\caption{Illustration diagram that shows the motivations of transferable contrastive learning. The training images should not only match their class semantics (discriminative property) but also have relatively high contrastive values with similar target classes (transfer property). `D' represents discriminative learning and `T' represents transfer learning.}
\label{fig:motivation}
\end{figure}

Although ZSL has made great progress in recent years, the strong assumption that the test images only come from the target classes is not realistic for practical applications. Therefore, generalized zero-shot learning (GZSL) \cite{chao2016empirical,Xian2017ZeroShotL} draws much attention recently, where test samples may come from either source or target classes. However, most existing ZSL approaches perform badly on GZSL task because they are easy to overfit the source classes, which indicates that they learn little knowledge about the target classes. These approaches learn the models only on the source categories and ignore the targets. Since the domain shift problem exists \cite{Fu2015TransductiveMZ}, the models learned on the source classes may not be suitable to the target classes, which results in overfitting the source categories in the GZSL task.

In order to tackle such problem, we propose to explicitly transfer the knowledge from the source classes to the target categories. The key problem for ZSL is that no labeled images are available for the target categories so we could not directly train the target image classifiers. An intuitive idea is to learn target classifiers from similar source images. For example, we could leverage the source images `horse' to learn the target class `zebra'. Based on this idea, we propose a novel transferable contrastive network for generalized zero-shot learning. It automatically contrasts the images with class semantics to judge whether they are consistent or not. Figure \ref{fig:motivation} shows the motivations of our approach, where two key properties for ZSL are considered in the contrastive learning process: discriminative property and transferable property. We maximize the contrastive values of images with corresponding class semantics and minimize the inconsistent ones among source classes thus to ensure that our model is discriminative enough to recognize different classes. Furthermore, to make the contrast transferable to the target classes, we utilize the class similarities to transfer knowledge from the source-class images to similar target classes. In this way, the model will be more robust to the target categories though no labeled target images are available to learn the model.

The main contributions of this paper are in two aspects. First, we propose a novel transferable contrastive network for GZSL, where a new network structure is designed for contrastive learning. Second, we consider both the discriminative property and transferable property in the contrastive learning procedure, where the discriminative property ensures to effectively discriminate different classes and the transferable property guarantees the robustness to the target classes. Experiments on five benchmark datasets show the superiority of the proposed approach.

%The rest of this paper is organized as follows: Section 2 reviews some related works for zero-shot learning. Section 3 introduces the contrastive network in detail and then Section 4 displays some experimental results and analysis. In the end, we give the concluding remarks in Section 5.

%-------------------------------------------------------------------------
\section{Related Work}

%-------------------------------------------------------------------------
\subsection{Semantic Information}

Semantic information is the key to ZSL. It builds up the relations between the source and target classes thus to enable knowledge transfer. Recently, the most widely used semantic information in ZSL is attributes \cite{lampert2009learning,farhadi2009describing} and word vectors \cite{mikolov2013distributed}. Attributes are general descriptions of objects. They are accurate but need human experts for definition and annotation. Word vectors are automatically learned from large numbers of text corpus which reduces human labor. However, there is much noise in the texts, which restricts their performance. In this paper, we use the attributes as the semantic information since they are more accurate to bridge the source and target classes.

%-------------------------------------------------------------------------
\subsection{Visual-Semantic Transformations}

Visual-semantic transformations establish relationships between the visual space and the semantic space. According to different projection directions, current ZSL approaches can be grouped into three types: visual to semantic embeddings, semantic to visual embeddings, latent space embeddings. We will introduce them in detail below.

\textbf{Visual to semantic embeddings.} These approaches learn the transformations from the visual space to the semantic space and perform image recognition in the semantic space. In the early age of ZSL, \cite{lampert2009learning,farhadi2009describing} propose to learn attribute classifiers to transfer knowledge from the source to the target classes. They train each attribute classifier independently, which is time-consuming. To tackle such problem, \cite{akata2013label,akata2015evaluation} consider all attributes as a whole and learn label embedding functions to maximize the compatibilities between images and corresponding class semantics. Furthermore, \cite{Norouzi2014ZeroShotLB} proposes to synthesize the semantic representations of test images by a convex combination of source-class semantics using the probability outputs of source classifiers. To learn more robust transformations, \cite{Morgado2017SemanticallyCR} proposes a deep neural network to combine attribute classifier learning and semantic label embedding.

\textbf{Semantic to visual embeddings.} These approaches learn the transformations from semantic space to the visual space and perform image recognition in the visual space, which can effectively tackle the hubness problem in ZSL \cite{dinu2014improving,shigeto2015ridge}. \cite{Changpinyo2017PredictingVE,Long2017ZeroshotLU,Zhang2017LearningAD} predict the visual samplers by learning embedding functions from the semantic space to the visual space. \cite{romera2015embarrassingly} adds some regularizers to learn the embedding function from class semantic to corresponding visual classifiers and \cite{wang2018zero} utilizes knowledge graphs to learn the same embedding functions. Some other works directly synthesize the target-class classifiers \cite{Changpinyo2016SynthesizedCF} or learn the target-class prototypes \cite{jiang2018learning} in the visual space by utilizing the class structure information. \cite{Kodirov2017SemanticAF,chen2018zero} exploit the auto-encoder framework to learn both the semantic to visual and visual to semantic embeddings simultaneously. Inspired by the generative adversarial networks, \cite{xian2018feature} generates the target-class samples in the feature space and directly learns the target classifiers.

\textbf{Latent space embedding.} These approaches encode the visual space and semantic space into a latent space for more effective image recognition. Since the predefined semantic information may be not discriminative enough to classify different classes, \cite{Zhang2015ZeroShotLV,Zhang2016ZeroShotLV} propose to use class similarities as the embedding space and \cite{Jiang2017LearningDL} proposes discriminative latent attributes for zero-shot recognition. Moreover, \cite{Bucher2016ImprovingSE} exploits metric learning techniques, where relative distance is utilized, to improve the embedding models. In order to learn robust visual-semantic transformations, \cite{frome2013devise,Socher2013ZeroShotLT,Reed2016LearningDR,Morgado2017SemanticallyCR} utilize deep neural networks to project the visual space and the semantic space into a common latent space and align the representations of the same class.

Our approach belongs to the latent space embedding, but there is a little difference. Traditional methods aim to minimize the distance of images and corresponding class semantics in the latent space for image recognition, while our approach fuses their information for contrastive learning.

%-------------------------------------------------------------------------
\subsection{Zero-Shot Recognition}

Zero-shot recognition is the last step for ZSL, most of which can be grouped into two categories. The distance-based approaches usually exploit the nearest neighbour approach to recognize the target-class samples \cite{akata2013label,Fu2015ZeroshotOR,Zhang2015ZeroShotLV,Jiang2017LearningDL} and the classifier-based approaches directly learn the visual classifiers to recognize the target-class images \cite{Changpinyo2016SynthesizedCF,xian2018feature}. Our approach utilizes contrastive values for image recognition.

%-------------------------------------------------------------------------
\subsection{Discussions about Relevant Works}

Most existing approaches ignore the target classes when learning the recognition model, so they are prone to overfitting the source classes in GZSL task. To tackle this problem, \cite{xian2018feature,Zhu2018AGA} leverage the semantic information of target classes to generate image features for training target classifiers. Although satisfactory performance has been achieved, it is difficult to train and use the generative models. While our approach is easy to learn. Moreover, it is complementary to such generative approaches. \cite{Liu2018GeneralizedZL} proposes a calibration network that calibrates the confidence of source classes and uncertainty of target classes. Different from it, we directly transfer knowledge to the target classes, which is more effective for GZSL. \cite{Fu2015TransductiveMZ} uses all the unlabeled target images to adjust the models in transductive ZSL settings. However, these images are often unavailable in practical conditions, so we perform the inductive ZSL task. \cite{Jiang2019AML} proposes adaptive metric learning to make the model suitable for the target classes. However, the linear model restricts its performance. Another relevant work is \cite{sung2018learning}, which also studies the relations between images and class semantics. Compared with \cite{sung2018learning}, we design a novel network structure for TCN. Moreover, we explicitly transfer knowledge from the source images to similar target classes, which makes our model more robust to the target categories.

%------------------------------------------------------------------------
\section{Approach}

\begin{figure*}[t]
\centering
\includegraphics[height=6.2cm]{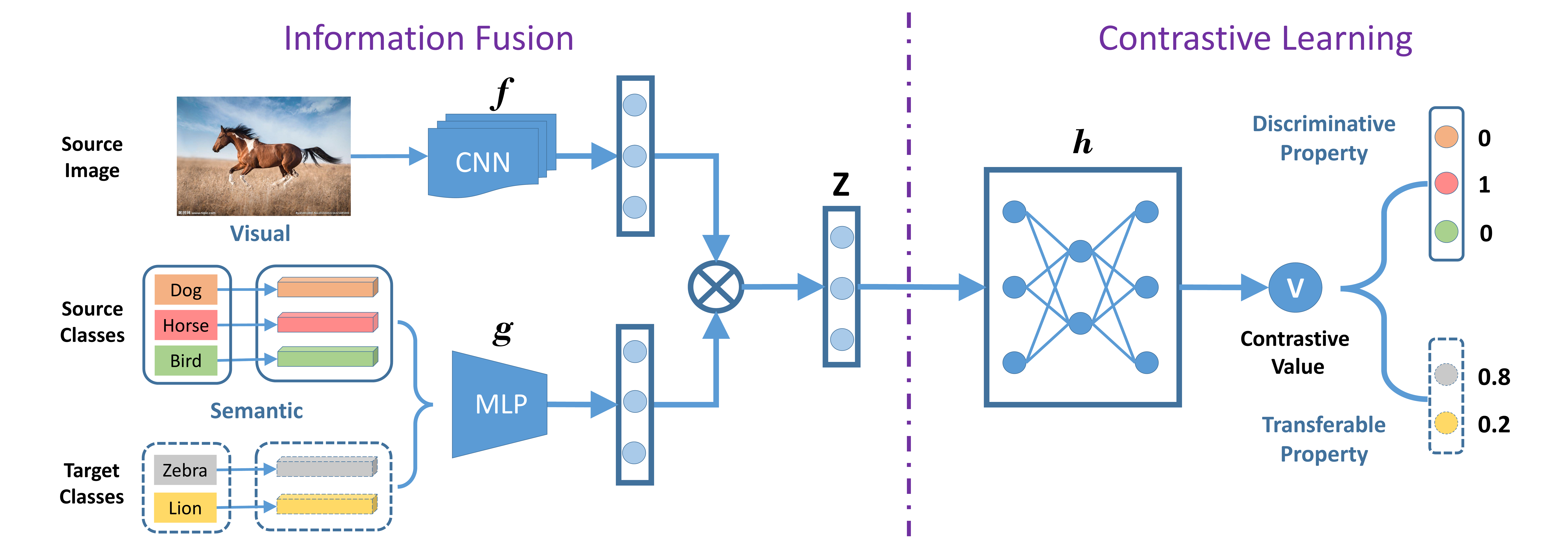}
\caption{The framework of transferable contrastive network. The information fusion module merges the image information with the class semantic information. The contrastive learning module automatically judges whether the fusion is consistent or not. `$\bigotimes$' denotes the element-wise product operation.}
\label{fig:framework}
\end{figure*}

The objective of our approach is learning how to contrast the image with the class semantics. Figure \ref{fig:framework} shows the general framework of the proposed transferable contrastive network (TCN). It contains two parts: information fusion and contrastive learning. Instead of computing the distance between images and class semantics using fixed metric for recognition, we fuse their information and learn a metric that automatically judges whether the fusions are consistent or not, where high contrastive values should be obtained between images and corresponding class semantics. In order to make the contrastive mechanism suitable to the target classes, we explicitly transfer knowledge from source images to similar target classes since no target images are available for training. More details will be described below.

%-------------------------------------------------------------------------
\subsection{Problem Settings}

In zero-shot learning, we are given $K$ source classes (denoted as $\mathcal{Y}^s$) and $L$ target classes (denoted as $\mathcal{Y}^t$), where the source and target classes are disjoint, \emph{i.e.} $\mathcal{Y}^{s} \cap \mathcal{Y}^{t} = \emptyset $. We use the index $\{1,...,K\}$ to represent the source classes and $\{K+1,...,K+L\}$ to represent the target classes. The source classes contain $N$ labeled images $\mathcal{D} = \{(\bm{x}_i,y_i)| \bm{x}_i \in \mathcal{X}, y_i \in \mathcal{Y}^s \}_{i=1}^{N}$, while no labeled images are available for the target classes. $\mathcal{X}$ represents the visual sample space. To build up the relations between the source and target classes, semantic information $\mathcal{A} = \{ \bm{a}_c\}_{c=1}^{K+L}$ is provided for each class $c \in \mathcal{Y}^{s} \cup \mathcal{Y}^{t}$. The goal of ZSL is to learn visual classifiers of target classes $f_{zsl} : \mathcal{X} \rightarrow \mathcal{Y}^{t}$ and the goal of GZSL is to learn more general visual classifiers of all classes $f_{gzsl} : \mathcal{X} \rightarrow \mathcal{Y}^{s} \cup \mathcal{Y}^{t}$.

%-------------------------------------------------------------------------
\subsection{Contrastive Network}

\textbf{Information Fusion.} An intuitive way of contrasting an image with one class is to fuse their information and judge how consistent the fusion is. Therefore, we first encode the images and class semantics into the same latent feature space to fuse their information. As is shown in Figure \ref{fig:framework}, we use two branches of neural network to encode the image and the class semantics into the same feature space respectively, where convolutional neural network (CNN) is utilized to encode the images and the multilayer perceptrons (MLP) is utilized to encode the class semantic information (attributes or word vectors). Then an element-wise product operation ($ \otimes$) is exploited to fuse the information from these two domains. Let $f(\bm{x}_i)$ denote the coding feature of the $i$th image and $g(\bm{a}_j)$ represent the coding feature of the $j$th class semantic, we can get the fused feature $\bm{z}_{ij}$ as:
\begin{equation}
  \bm{z}_{ij} = f(\bm{x}_i) \otimes g(\bm{a}_j)
  \label{equ:fusion}
\end{equation}
where $\bm{a}_j$ is the class semantic of the $j$th class. Then we can feed $\bm{z}_{ij}$ to the next stage to judge how well the image $i$ is consistent with class $j$.

\textbf{Contrastive Learning.} Different from previous approaches that use fixed distance, such as Euclidean distance or cosine distance, to compute the similarities between images and classes for image recognition, we design a contrastive network that automatically judges how well the image is consistent with a specific class. Let $v_{ij}$ denote the contrastive value between image $i$ and class $j$, we can obtain it from the fused feature $\bm{z}_{ij}$ as:
\begin{equation}
  v_{ij} = h(\bm{z}_{ij})
  \label{equ:contrast}
\end{equation}
where $h$ is the contrastive learning function.

In the contrastive learning phase, we should consider two characters: discriminative property and transferable property. Discriminative property indicates that the contrastive model should be discriminative enough to classify different classes. Transferable property means that the contrastive model should be generalized to the target classes.

In order to enable the discriminative property, we utilize the semantic information of source classes as supervision, where the contrastive values of consistent fusions are maximized and those of inconsistent ones are minimized. The loss function can be formulated by the cross-entropy loss:
\begin{equation}
  \mathcal{L}_D = - \sum_{i=1}^{N} \sum_{j=1}^{K} m_{ij} \log v_{ij} + (1-m_{ij}) \log(1-v_{ij})
  \label{equ:discriminative}
\end{equation}
where $m_{ij}$ is a class indicator. Let $y_i$ be the class label for the $i$th image, then $m_{ij}$ can be obtained by:
\begin{equation}
  m_{ij} = \left\{\begin{matrix}
  1, \quad y_i = j
\\
0, \quad y_i \neq j
\end{matrix}\right.
  \label{equ:indicator}
\end{equation}

The goal of ZSL is to recognize the target classes. If we only use the source classes in the contrastive learning phase, it is easy to overfit and the model would be less transferable to the target classes. This is the problem that exists in most ZSL approaches. Unfortunately, we don't have labeled target images to take part in the contrastive learning process. To tackle such problem, we explicitly transfer knowledge from source images to the target classes by class similarities. In other words, the source images could also be utilized to learn similar target classes. Let $s_{kj}$ denote the similarity of source class $k$ (k=1,...,K) to target class $j$ (j=K+1,...,K+L) and then the loss function for transferable property is formulated as:
\begin{equation}
  \mathcal{L}_T = - \sum_{i=1}^{N} \sum_{j=K+1}^{K+L} s_{y_ij} \log v_{ij} + (1-s_{y_ij}) \log(1-v_{ij})
  \label{equ:transfer}
\end{equation}

To summarize, our full loss function is:
\begin{equation}
  \mathcal{L} = \mathcal{L}_D + \alpha \mathcal{L}_T
  \label{equ:full}
\end{equation}
where $\alpha$ is a parameter that controls the relative importance of discriminative property and transferable property.

%-------------------------------------------------------------------------
\subsection{Class Similarity}

In order to accomplish the contrastive learning approach proposed above, the similarities between the source and target classes should be obtained. Inspired by the sparse coding approach, we utilize the target classes to reconstruct a source class and the reconstruction coefficients are viewed as the similarity of the source class to the target classes. The objective function is:
\begin{equation}
  \bm{s}_{k} = \arg \min \limits_{\bm{s}_{k}} ||\bm{a}_k - \sum_{j=K+1}^{K+L} \bm{a}_js_{kj}||_{2}^{2} + \beta||\bm{s}_{k}||_{2}
  \label{equ:class_sim}
\end{equation}
where $\bm{a}_k$ is the semantic information of class $k$ and $s_{kj}$ is the $j$th element of $\bm{s}_{k}$, which denotes the similarities of source class $k$ to target class $j$. $\beta$ is the regularization parameter. Then we normalize the similarity by
\begin{equation}
  s_{kj} = \frac{s_{kj}}{\sum_{j=K+1}^{K+L}s_{kj}}
  \label{equ:norm}
\end{equation}

%-------------------------------------------------------------------------
\subsection{Zero-Shot Recognition}

We conduct zero-shot recognition by comparing the contrastive values of one image with all the class semantics.

For ZSL, we classify one image to the class which has the largest contrastive value among target classes, which can be formulated as:
\begin{equation}
  P_{zsl}(x_i) = \max_j \{v_{ij}\}_{j=K+1}^{K+L}
  \label{equ:zsl}
\end{equation}

For GZSL, we classify one image to the class which has the largest contrastive value among all classes, which can be formulated as:
\begin{equation}
  P_{gzsl}(x_i) = \max_j \{v_{ij}\}_{j=1}^{K+L}
  \label{equ:gzsl}
\end{equation}

%------------------------------------------------------------------------
\section{Experiment}
%-------------------------------------------------------------------------
\subsection{Datasets and Settings}

\begin{table}[t]
\begin{small}
\begin{center}
\begin{tabular}{|l|c|c|c|c|}
\hline
\textbf{Dataset}     &   \quad \emph{Img}    &  \quad \emph{Attr}   &  \quad \emph{Source}  &  \quad \emph{Target} \\
\hline
\hline
\textbf{APY} \cite{farhadi2009describing}       &  \quad 15,339   &  \quad 64   &  \quad 15 + 5     &  \quad 12   \\
\textbf{AWA1} \cite{lampert2009learning}        &  \quad 30,475   &  \quad 85   &  \quad 27 + 13    &  \quad 10   \\
\textbf{AWA2} \cite{Xian2018ZeroShotL}          &  \quad 37,322   &  \quad 85   &  \quad 27 + 13    &  \quad 10   \\
\textbf{CUB} \cite{wah2011caltech}              &  \quad 11,788   &  \quad 312  &  \quad 100 + 50   &  \quad 50   \\
\textbf{SUN} \cite{patterson2014sun}            &  \quad 14,340   &  \quad 102  &  \quad 580 + 65   &  \quad 72 \\
\hline
\end{tabular}
\end{center}
\end{small}
\caption{
Statistics for attribute datasets: APY , AWA1, AWA2, CUB and SUN in terms of image numbers (\emph{Img}), attribute numbers (\emph{Attr}), training + validation source class numbers (\emph{Source}) and target class numbers (\emph{Target}).
}
\label{table:database}
\end{table}

We conduct experiments on five widely used ZSL datasets: APY \cite{farhadi2009describing}, AWA (2 versions AWA1 \cite{lampert2009learning} and AWA2 \cite{Xian2018ZeroShotL}), CUB \cite{wah2011caltech}, SUN \cite{patterson2014sun}. APY is a small-scale coarse-grained dataset with 64 attributes, which contains 20 object classes of aPascal and 12 object classes of aYahoo. AWA1 is a medium-scale animal dataset which contains 50 animal classes with 85 attributes annotated. AWA2 is collected by \cite{Xian2018ZeroShotL}, which has the same classes as AWA1. CUB is a fine-grained and medium-scale dataset, which contains 200 different types of birds annotated with 312 attributes. SUN is a medium-scale dataset containing 717 types of scenes where 102 attributes are annotated. In order to make fair comparisons with other approaches, we conduct our experiment on the more reasonable pure ZSL settings recently proposed by \cite{Xian2018ZeroShotL}. The details of each dataset and class splits for source and target classes are shown in Table \ref{table:database}.

\textbf{Implementation Details.} We extract the image features $f(\bm{x})$ by the ResNet101 model \cite{Kaiming2016} and use class attributes as the semantic information. The class semantic transformation $g(\bm{a})$ is implemented by a two-layer fully connected neural network, where the hidden layer dimension is set to 1024 and the output size is 2048. The contrastive learning $h(\bm{z})$ is also implemented by the fully connected neural network, where the hidden dimension is 1024 and the output size is 1. We use Leaky ReLU as the nonlinear activation function for all the hidden layers and sigmoid function for the last layer \footnote{Source code is available at \emph{http://vipl.ict.ac.cn/resources/codes}.}. The hyperparameter $\alpha$ is fine-tuned in the range [0.001, 0.01, 0.1, 1] by the validation set.

%-------------------------------------------------------------------------
\subsection{Performance on ZSL and GZSL}
\begin{table}[t]
\begin{small}
\begin{center}
\begin{tabular}{|l|c|c|c|c|c|}
\hline
\textbf{Method}     &  \textbf{APY }    &  \textbf{AWA1}   &  \textbf{AWA2}  & \textbf{CUB }  &  \textbf{SUN} \\
\hline
\hline
DAP \cite{lampert2009learning}              &   33.8   &   44.1   &   46.1   &   40.0   &   39.9   \\
IAP \cite{lampert2009learning}              &   36.6   &   35.9   &   35.9   &   24.0   &   19.4   \\
CONSE \cite{Norouzi2014ZeroShotLB}          &   26.9   &   45.6   &   44.5   &   34.3   &   38.8   \\
CMT \cite{Socher2013ZeroShotLT}             &   28.0   &   39.5   &   37.9   &   34.6   &   39.9 \\
SSE \cite{Zhang2015ZeroShotLV}              &   34.0   &   60.1   &   61.0   &   43.9   &   51.5   \\
LATEM \cite{Xian2016LatentEF}               &   35.2   &   55.1   &   55.8   &   49.3   &   55.3   \\
ALE \cite{akata2013label}                   &   39.7   &   59.9   &   62.5   &   54.9   &   58.1   \\
DEVISE \cite{frome2013devise}               &   39.8   &   54.2   &   59.7   &   52.0   &   56.5 \\
SJE \cite{akata2015evaluation}              &   32.9   &   65.6   &   61.9   &   53.9   &   53.7   \\
EZSL \cite{romera2015embarrassingly}        &   38.3   &   58.2   &   58.6   &   53.9   &   54.5   \\
SYNC \cite{Changpinyo2016SynthesizedCF}     &   23.9   &   54.0   &   46.6   &   55.6   &   56.3   \\
SAE \cite{Kodirov2017SemanticAF}            &   8.3    &   53.0   &   54.1   &   33.3   &   40.3 \\
CDL \cite{jiang2018learning}                &  43.0   &   69.9   &   -      &   54.5   &   \textbf{63.6} \\
RNet \cite{sung2018learning}                &   -      &   68.2   &   64.2   &   55.6   &   - \\
FGN \cite{xian2018feature}               &   -      &   68.2   &   -   &   57.3   & 60.8 \\
GAZSL \cite{Zhu2018AGA}                		&   41.1   &   68.2   &   70.2   &   55.8   &  61.3 \\
DCN \cite{Liu2018GeneralizedZL}             &    \textbf{43.6}      &  65.2   &   -   &   56.2   & 61.8 \\
\hline
TCN (ours)                                 &   38.9   &   \textbf{70.3}   &   \textbf{71.2}   &   \textbf{59.5}   &   61.5 \\
\hline
\end{tabular}
\end{center}
\end{small}
\caption{
Zero-shot recognition results on APY, AWA1, AWA2, CUB and SUN (\%). `-' denotes that the results are not reported.
}
\label{table:compare}
\end{table}

\begin{table*}[t]
\small
\begin{center}
\begin{tabular}{|p{2cm}|p{0.5cm}p{0.5cm}p{0.5cm}|p{0.5cm}p{0.5cm}p{0.5cm}|p{0.5cm}p{0.5cm}p{0.5cm}|p{0.5cm}p{0.5cm}p{0.5cm}|p{0.5cm}p{0.5cm}p{0.5cm}|}
\hline
\multirow{2}{*}{Method}&
    \multicolumn{3}{c|}{APY}&\multicolumn{3}{c|}{AWA1}&\multicolumn{3}{c|}{AWA2}&\multicolumn{3}{c|}{CUB}&\multicolumn{3}{c|}{SUN}\cr
    &ts & tr & H &ts & tr & H  &ts & tr & H  &ts & tr & H  &ts & tr & H  \cr
\hline
DAP \cite{lampert2009learning}            &4.8   &78.3   &9.0   &0.0   &88.7   &0.0    &0.0   &84.7   &0.0    &1.7    &67.9   &3.3    &4.2    &25.1   &7.2 \\
IAP \cite{lampert2009learning}            &5.7   &65.6   &10.4  &2.1   &78.2   &4.1    &0.9   &87.6   &1.8    &0.2    &72.8   &0.4    &1.0    &37.8   &1.8 \\
CONSE \cite{Norouzi2014ZeroShotLB}        &0.0   &91.2   &0.0   &0.4   &88.6   &0.8    &0.5   &90.6   &1.0    &1.6    &72.2   &3.1    &6.8    &39.9   &11.6 \\
CMT \cite{Socher2013ZeroShotLT}           &1.4   &85.2   &2.8   &0.9   &87.6   &1.8    &0.5   &90.0   &1.0    &7.2    &49.8   &12.6   &8.1    &21.8   &11.8 \\
SSE \cite{Zhang2015ZeroShotLV}            &0.2   &78.9   &0.4   &7.0   &80.5   &12.9   &8.1   &82.5   &14.8   &8.5    &46.9   &14.4   &2.1    &36.4   &4.0 \\
LATEM \cite{Xian2016LatentEF}             &0.1   &73.0   &0.2   &7.3   &71.7   &13.3   &11.5  &77.3   &20.0   &15.2   &57.3   &24.0   &14.7   &28.8   &19.5 \\
ALE \cite{akata2013label}                 &4.6   &73.7   &8.7   &16.8  &76.1   &27.5   &14.0  &81.8   &23.9   &23.7   &62.8   &34.4   &21.8   &33.1   &26.3 \\
DEVISE \cite{frome2013devise}             &4.9   &76.9   &9.2   &13.4  &68.7   &22.4   &17.1  &74.7   &27.8   &23.8   &53.0   &32.8   &16.9   &27.4   &20.9 \\
SJE \cite{akata2015evaluation}            &3.7   &55.7   &6.9   &11.3  &74.6   &19.6   &8.0   &73.9   &14.4   &23.5   &59.2   &33.6   &14.1   &30.5   &19.8 \\
EZSL \cite{romera2015embarrassingly}      &2.4   &70.1   &4.6   &6.6   &75.6   &12.1   &5.9   &77.8   &11.0   &12.6   &63.8   &21.0   &11.0   &27.9   &15.8 \\
SYNC \cite{Changpinyo2016SynthesizedCF}   &7.4   &66.3   &13.3  &8.9   &87.3   &16.2   &10.0  &90.5   &18.0   &11.5   &70.9   &19.8   &7.9    &43.3   &13.4 \\
SAE \cite{Kodirov2017SemanticAF}          &0.4   &80.9   &0.9   &1.8   &77.1   &3.5    &1.1   &82.2   &2.2    &7.8    &54.0   &13.6   &8.8    &18.0   &11.8 \\
CDL \cite{jiang2018learning}              &19.8  &48.6   &28.1  &28.1  &73.5   &40.6   &-     &-      &-      &23.5   &55.2   &32.9   &21.5   &34.7   &26.5 \\
RNet \cite{sung2018learning}              &-     &-      &-     &31.4  &91.3   &46.7   &30.0  &93.4   &45.3   &38.1   &61.4   &47.0   &-      &-      &-    \\
FGN \cite{xian2018feature}             &-     &-      &-     &57.9  &61.4   &59.6   &-     &-      &-   &43.7   &57.7   &49.7   &42.6    &36.6 &\textbf{39.4}    \\
GAZSL \cite{Zhu2018AGA}               &14.2  &78.6   &24.0  &29.6  &84.2   &43.8   &35.4     &86.9      &50.3      &31.7   &61.3   &41.8   &22.1 &39.3   &28.3 \\
DCN \cite{Liu2018GeneralizedZL}        &14.2  &75.0   &23.9  &25.5  &84.2   &39.1   &-     &-      &-      &28.4   &60.7   &38.7   &25.5   &37.0 &30.2 \\
\hline
TCN (ours)                               &24.1  &64.0   &\textbf{35.1}  &49.4  &76.5   &\textbf{60.0}   &61.2  &65.8   &\textbf{63.4}   &52.6 &52.0   &\textbf{52.3}   &31.2   &37.3   &34.0 \\
\hline
\end{tabular}
\end{center}
\caption{
GZSL results on APY, AWA1, AWA2, CUB and SUN. ts = Top-1 accuracy of the target classes, tr = Top-1 accuracy of the source classes, H = harmonic mean. We measure average per-class top-1 accuracy in \%. `-' represents that the results are not reported.
}
\label{table:compare2}
\end{table*}

To demonstrate the effectiveness of the transferable contrastive network, we compare our approach with several state-of-the-art approaches. Table \ref{table:compare} shows the comparison results of ZSL, where the performance is evaluated by the average per-class top-1 accuracy. It can be seen that our approach achieves the best performance on three datasets and is comparable to the best approach on SUN, which indicates that transferable contrastive network can make good knowledge transfer to the target classes. Our approach is effective to perform fine-grained recognition, as can be seen by the good performance on CUB. We owe the success to two aspects. First, the discriminative property of contrastive learning ensures the contrastive network to effectively discriminate the fine-grained classes. Second, the fine-grained images are more effective to transfer the knowledge since the classes are similar, which makes our model more robust to the target classes. A little lower performance is obtained on APY probably due to the weak relations between the source and target classes. APY is a small-scale coarse-grained dataset, where the categories are very different. Therefore, the relations between source and target classes are weak. That's why most approaches could not perform well on this simple dataset. Since we utilize the class similarities to transfer the knowledge, our model may be influenced by the weak relations.

We argue that traditional approaches usually tend to overfit the source classes since they ignore the target in the model learning process, which will result in the projection domain shift problem. While TCN could alleviate this problem since our model explicitly transfers the knowledge. To demonstrate this viewpoint, we perform GZSL task on these datasets. Table \ref{table:compare2} shows the comparison results, where `ts' is average per-class top-1 accuracy of target classes and `tr' is the same evaluation results on source classes. `H' is the harmonic mean that evaluates the total performance. It can be seen that most approaches achieve very high performance on the source classes and extremely low performance on the target classes, which indicates that these approaches learn little knowledge about the target classes. Compared with the results in Table \ref{table:compare}, the performance of target classes drops greatly for GZSL because most target-class images are recognized as source classes. This indicates that previous approaches are easy to overfit the source classes. While TCN can effectively alleviate the overfitting problem, as can be seen by the more balanced performance on source and target classes for our approach. We owe the success to the transferable property of the contrastive network, which makes our model more robust to recognize the target images. Although the generative approaches \cite{xian2018feature,Zhu2018AGA} are also very effective in GZSL, they need to learn the complicated generative models. While our approach is very simple to learn. Moreover, our approach is well complementary to the generative approaches since the generated features can also be utilized to learn our model. Some other approaches \cite{jiang2018learning,Liu2018GeneralizedZL} also adapt the models to the target classes. Compared with them, our approach is more effective.

We also tried other information fusion approaches and more details are shown in the supplementary materials.

%-------------------------------------------------------------------------
\subsection{Importance of Knowledge Transfer}
\begin{table}[t]
\small
\begin{center}
\begin{tabular}{|c|c|c|c|c|c|}
\hline
\multirow{2}{*}{\textbf{Dataset}}&  \multirow{2}{*}{\textbf{Method}}&    \multirow{2}{*}{\textbf{ZSL}}&      \multicolumn{3}{c|}{\textbf{GZSL}}\cr
\cline{4-6}
    & {} & {} &\textbf{ts}   &  \textbf{tr}  & \textbf{H} \\
\hline
\multirow{2}{*}{APY}   & Base  &37.52  & 5.50  & 77.78  & 10.28 \cr  & TCN & 38.93  & 24.13  & 64.00  & 35.05\cr
\hline
\multirow{2}{*}{AWA1}  & Base  &70.15  & 9.22  & 64.78  & 16.14 \cr  & TCN  & 70.34  & 49.40  & 76.48  & 60.03\cr
\hline
\multirow{2}{*}{AWA2}  & Base  &68.48  & 9.32  & 54.23  & 15.91 \cr  & TCN  & 71.18  & 61.20  & 65.83  & 63.43\cr
\hline
\multirow{2}{*}{CUB}   & Base  &56.62  & 24.70 & 64.90  & 37.84 \cr  & TCN  & 59.54  & 52.58  & 52.03  & 52.30\cr
\hline
\multirow{2}{*}{SUN}   & Base  &61.04  & 21.94 & 38.64  & 27.99 \cr  & TCN  & 61.53  & 31.18  & 37.29  & 33.96\cr
\hline
\end{tabular}
\end{center}
\caption{
Comparison with the baseline approach where the knowledge transfer item ($\mathcal{L}_T$) is removed. `Base' represents the baseline approach. `TCN' is our approach. `ZSL' is the accuracy of zero-shot recognition. `ts', `tr' and `H' are the target-class accuracy, source-class accuracy and harmonic mean in GZSL.
}
\label{table:compare3}
\end{table}

Explicit knowledge transfer is an important part of our framework. It is intuitive that similar objects should play a more important role in transfer learning. Therefore, we use class similarities to explicitly transfer the knowledge from source images to similar target classes. In this way, our model will be more robust to the target classes. Moreover, it should also have the ability to prevent the model from overfitting the source classes. To demonstrate these assumptions, we compare our approach with the basic model, where the knowledge transfer term ($\mathcal{L}_T$) is removed. Table \ref{table:compare3} shows the recognition results. Although only small improvements are achieved for ZSL, the improvements for GZSL are significant. This phenomenon demonstrates that explicit knowledge transfer can effectively tackle the overfitting problem, which enables the model to learn the knowledge about the target classes.

%\begin{figure}[t]
%\begin{minipage}[t]{1\linewidth}
%\centering
%\includegraphics[height=4.2cm]{parameter_awa1}\\
%\includegraphics[height=4.2cm]{parameter_cub}
%\caption{The recognition results on AWA1 and CUB with different value of $\alpha$. `ZSL' is the accuracy of zero-shot recognition. `ts', `tr' and `'H' are the unseen-class accuracy, seen-class accuracy and harmonic mean in GZSL. }
%\label{fig:parameter}
%\end{minipage}%
%\end{figure}

\begin{figure}[t]
\centering
\includegraphics[height=4.2cm]{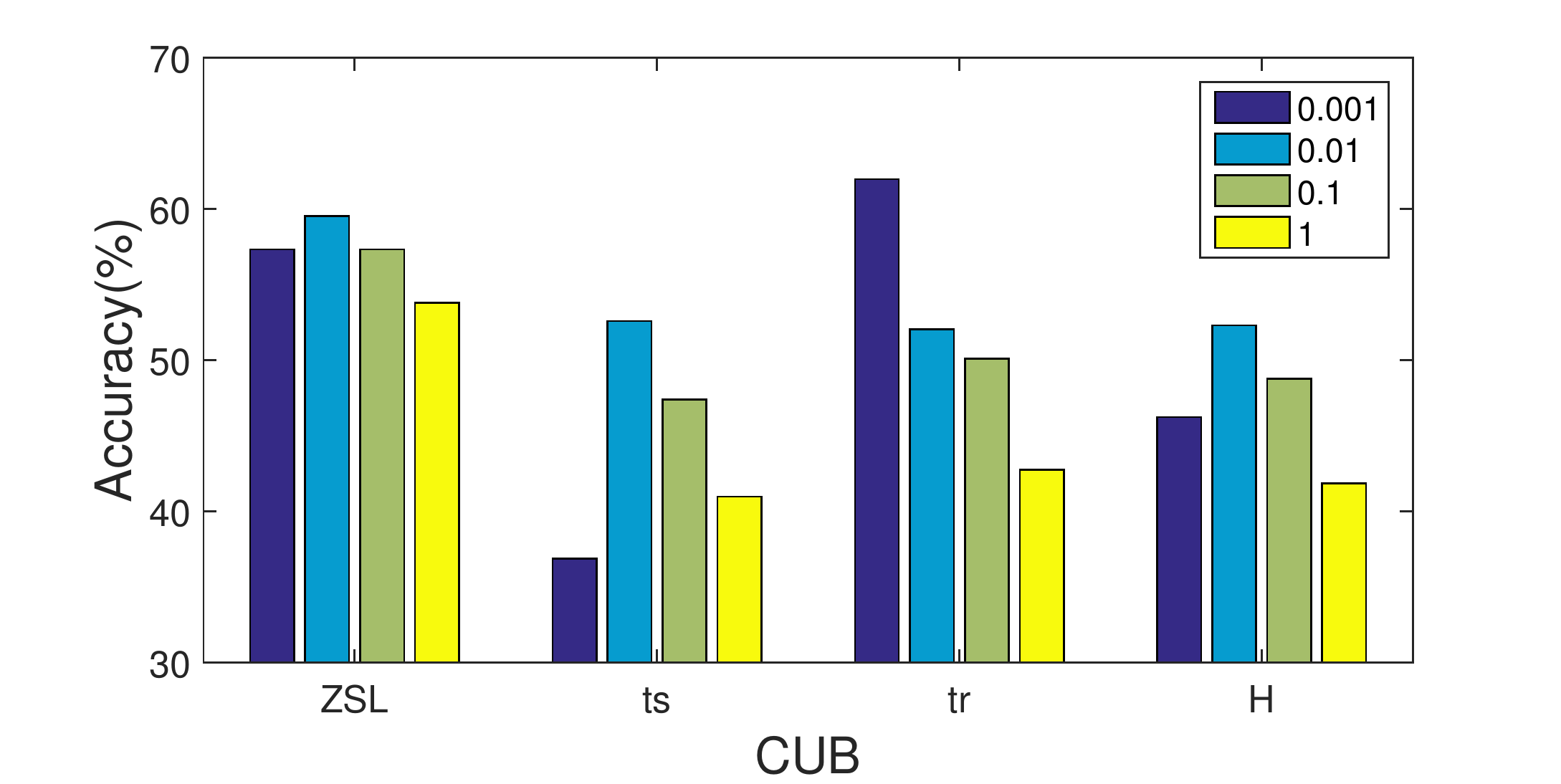}
\caption{The recognition results on CUB with different value of $\alpha$. `ZSL' is the accuracy of zero-shot recognition. `ts', `tr' and `H' are the target-class accuracy, source-class accuracy and harmonic mean in GZSL. }
\label{fig:parameter}
\end{figure}

Another factor that deserves to be explored is how important the knowledge transfer is. Therefore, we analyze the influence of parameter $\alpha$ to our model and the recognition results on CUB are shown in Figure \ref{fig:parameter}. It can be seen that TCN achieves its best performance when $\alpha$ equals to 0.01. We can infer that $\alpha$ should be small in order to get good performance. This may be caused by two reasons. First, the class similarities are fuzzy measures and there is no accurate definitions. Second, the source images do not absolutely match with the target classes. When $\alpha$ increases, the performance of source classes drops, as can be seen by the results of `tr', because the model pays more attention to the target classes and neglects the accurate source classes. Since the loss on the source classes ensures the discriminative property of contrastive learning and the loss on the target classes guarantees the transferable property, we must balance these terms to obtain a robust recognition model.

\subsection{Visualization of Class Similarities}
\begin{figure}[t]
\centering
\includegraphics[height=6cm]{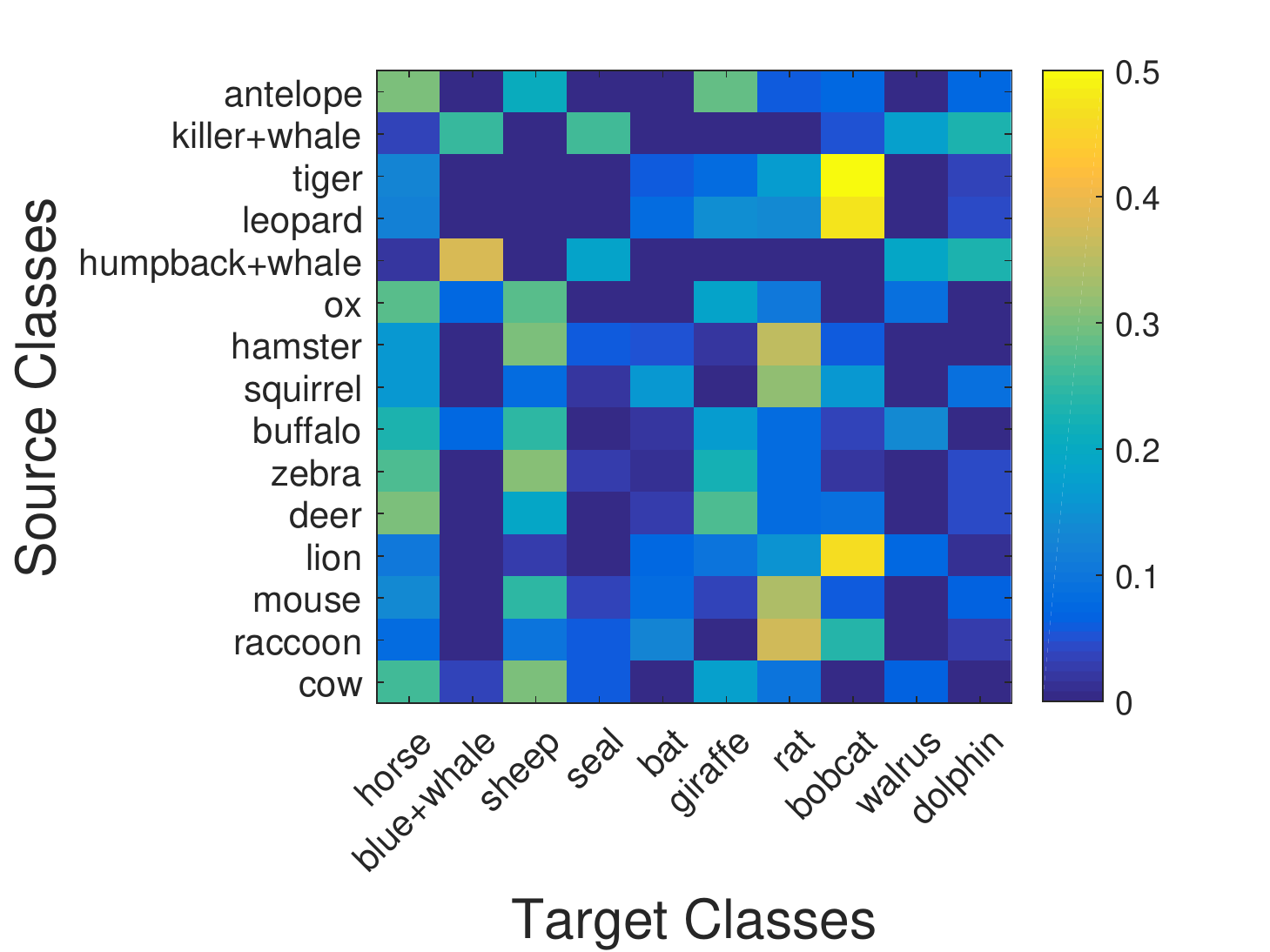}
\caption{The class similarities in AWA1, where 15 source classes are selected. Each row represents the similarities of one source class to the target classes.}
\label{fig:sim_awa}
\end{figure}

The transferable property of our approach is accomplished by leveraging the class similarities to make knowledge transfer in the model learning process. To see what knowledge has been transferred, we show the class similarities of AWA1 in Figure \ref{fig:sim_awa}. Because of space constraints, we select 15 source classes and visualize their similarities to the target classes. It can be figured out that \emph{leopard} is similar to \emph{bobcat} so the training samples of \emph{leopard} can also be utilized to learn the target class \emph{bobcat} in the training phase, thus to enable knowledge transfer. It effectively tackles the problem that no training images are available for the target classes. Through such explicit knowledge transfer, our model would be more robust to the target. Other class similarities, \emph{i.e.} \emph{killer+whale} is similar to \emph{blue+whale}, \emph{seal}, \emph{walrus} and \emph{dolphin}, are also useful knowledge to transfer in the contrastive learning process.

\begin{figure}[t]
\centering
\includegraphics[height=6cm]{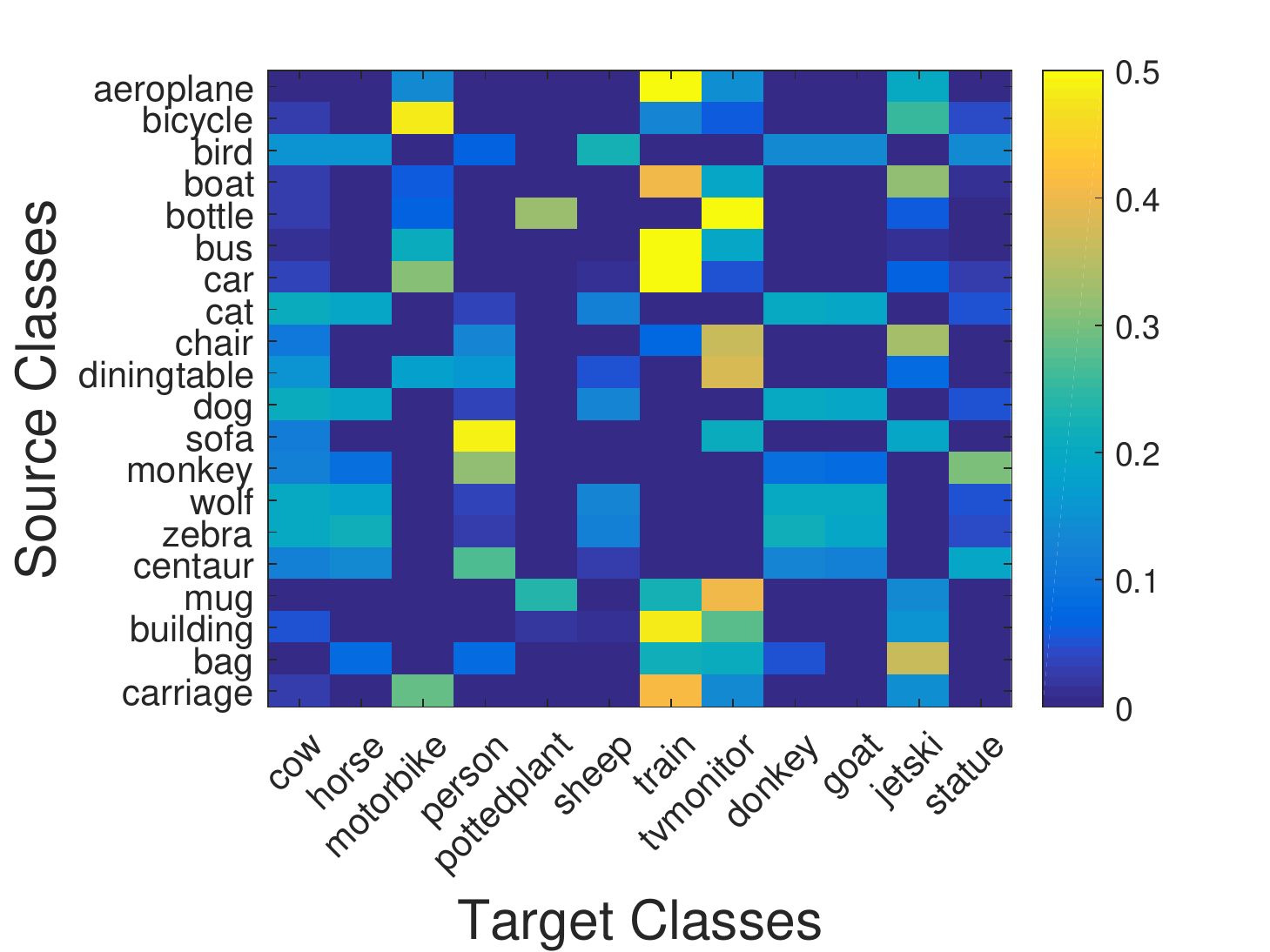}
\caption{The class similarities in APY, where each row shows the similarities of one source class to the target classes.}
\label{fig:sim_apy}
\end{figure}

\begin{figure*}[t]
\centering
\includegraphics[height=7cm]{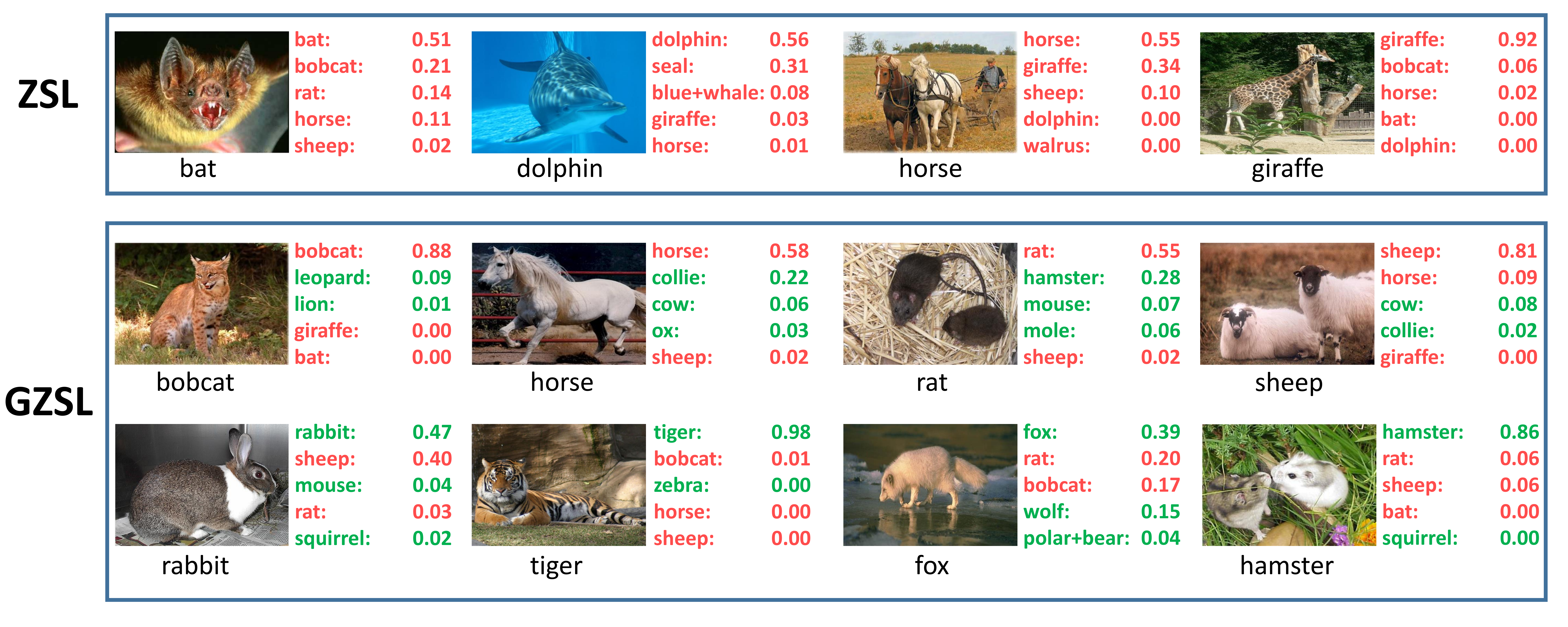}
\caption{The normalized contrastive values of some test samples obtained on AWA1, where five most similar classes are shown. The source classes are marked with green and the target classes are marked with red. The first row in GZSL shows the target-class samples and the second shows the source-class samples.}
\label{fig:contrast_visual}
\end{figure*}

The foundation on which our approach works well is that reasonable class similarities are obtained for knowledge transfer. However, the class similarities may be very rough for some coarse-grained dataset, such as APY, so it becomes difficult to transfer knowledge from source classes to the target classes. That is why low zero-shot recognition accuracy is obtained on APY for all approaches, as can be seen from Table \ref{table:compare}. To make it intuitive, we show the class similarities for APY in Figure \ref{fig:sim_apy}. It can be figured out that the relations between source and target classes are less reliable. For example, among the target classes, the most similar one to the source class \emph{building} is the \emph{train}. However, buildings and trains are very different in reality. Therefore, using the training images of \emph{building} to learn the target \emph{train} would degrade our model. This may be the reason why TCN achieves lower performance than the state-of-the-art approach on APY. Although some incomprehensible similarities exist, there are also some useful relations, \emph{i.e.}  \emph{bicycle} is similar to \emph{motorbike} and \emph{bus} is similar to \emph{train}, which ensures the relative good performance of our approach.

%----------------------------------------------------------------------
\subsection{Visualization of Contrastive Values}
Different from the visual-semantic embedding approaches that use fixed distance to conduct zero-shot recognition, our transferable contrastive network automatically contrasts one image with every class and outputs the contrastive values for image recognition. Figure \ref{fig:contrast_visual} shows the contrastive values of some test samples obtained on AWA1. In order to make it intuitive, we normalize the contrastive values and show five most similar classes, where the target classes are marked with red and the source classes are marked with green. We can figure out that most images are consistent with their corresponding classes and dissimilar with other classes. For ZSL, we recognize the test samples among the target classes. As can be seen, the image `giraffe' has high contrastive value with its class and has low contrastive values with other ones. For GZSL, we recognize the test samples among all classes. Although we only have source-class images for training, our model can effectively recognize the target-class samples in the test procedure. For example, `bobcat' is effectively discriminated with source-class \emph{leopard} in GZSL task though these two classes are very similar. We owe this success to the explicit knowledge transfer by the class similarities. It prevents our model from overfitting the source classes and ensures the transferable ability to target classes, thus the target-class images would be effectively recognized when they are encountered. Moreover, one image may also have relatively high contrastive values with similar classes. For example, `rat' has relative strong activations on \emph{hamster}. This shows that TCN is not only discriminative enough to classify different classes but also transferable to novel classes.

%-----------------------------------------, -----------------------------
\section{Conclusion}

In this paper, we propose a novel transferable contrastive network for generalized zero-shot learning. It automatically contrasts the images with the class semantics to judge how consistent they are. We consider two key properties in contrastive learning, where the discriminative property ensures the contrastive network to effectively classify different classes and the transferable property makes the contrastive network more robust to the target classes. By explicitly transferring knowledge from source images to similar target classes, our approach can effectively tackle the problem of overfitting the source classes in GZSL task. Extensive experiments on five benchmark datasets show the superiority of the proposed approach.

~\\
\noindent \textbf{Acknowledgements.} This work is partially supported by 973 Program under contract No. 2015CB351802, Natural Science Foundation of China under contracts Nos.61390511, 61772500, CAS Frontier Science Key Research Project No. QYZDJ-SSWJSC009, and Youth Innovation Promotion Association No. 2015085.

{\small
\bibliographystyle{ieee_fullname}
\bibliography{egbib}
}

\end{document}